\documentclass[journal,twoside]{IEEEtran}
\usepackage{amsmath}
\usepackage{amsfonts}
\usepackage{lipsum}
\usepackage{float}
\usepackage[hidelinks]{hyperref}
\usepackage{siunitx}
\usepackage{xcolor}
\usepackage{color}
\usepackage{graphicx}
\usepackage{amssymb}
\usepackage{textcomp}
\usepackage{subfig}
\usepackage{multirow}
\usepackage{booktabs}
\usepackage{colortbl}
\usepackage{algorithm}
\usepackage{algpseudocode}
\graphicspath{{images/}}
\definecolor{mygray}{gray}{.9}

\begin{document}
\title{RAZE: Region Guided Self-Supervised Gaze Representation Learning}

\author{Neeru Dubey*, Shreya Ghosh* and Abhinav Dhall 
\IEEEcompsocitemizethanks{
\IEEEcompsocthanksitem * \textit{The first two authors contributed equally to this work.}
\IEEEcompsocthanksitem N. Dubey is with Indian Institute of Technology Ropar, India. (E-mail: neerudubey@iitrpr.ac.in)
\IEEEcompsocthanksitem S. Ghosh is with Monash University. (Email: shreya.ghosh@monash.edu)
\IEEEcompsocthanksitem A. Dhall is with Monash University and Indian Institute of Technology Ropar, India. (E-mail: abhinav.dhall@monash.edu)
}
\thanks{\textbf{Project Page:} \href{https://sites.google.com/view/eyegazeproject/home}{https://sites.google.com/view/eyegazeproject/home}}
		
}

\markboth{}
{D\MakeLowercase{\textit{ubey et al.}}: RAZE: Region Guided Self-Supervised Gaze Representation Learning}

\maketitle

\begin{abstract}
Automatic eye gaze estimation is an important problem in vision based assistive technology with use cases in different emerging topics such as augmented reality, virtual reality and human-computer interaction. Over the past few years, there has been an increasing interest in unsupervised and self-supervised learning paradigms as it overcomes the requirement of large scale annotated data. In this paper, we propose RAZE, a Region guided self-supervised gAZE representation learning framework which leverage from non-annotated facial image data. RAZE learns gaze representation via auxiliary supervision i.e. pseudo-gaze zone classification where the objective is to classify visual field into different gaze zones (i.e. left, right and center) by leveraging the relative position of pupil-centers. Thus, we automatically annotate pseudo gaze zone labels of 154K web-crawled images and learn feature representations via `Ize-Net' framework. `Ize-Net' is a capsule layer based CNN architecture which can efficiently capture rich eye representation. The discriminative behaviour of the feature representation is evaluated on four benchmark datasets: \textit{CAVE, TabletGaze, MPII and RT-GENE}. Additionally, we evaluate the generalizability of the proposed network on two other downstream task (i.e. driver gaze estimation and visual attention estimation) which demonstrate the effectiveness of the learnt eye gaze representation.
	
\end{abstract}


\IEEEpeerreviewmaketitle

\section{Introduction}

\IEEEPARstart{T}{he} gaze estimation is a process of identifying the line-of-sight of the pupils at a particular instant. Eye gaze provides an important information about human visual attention and cognitive process~\cite{mason2004look,sun2014toward,wang2017deep}. It has a wide range of interactive applications including human-robot interaction~\cite{ghosh2018speech,zhou2022rfnet,sharma2021gaze}, student engagement detection~\cite{kaur2018prediction}, video games~\cite{barr2007video,cheng2013gaze}, driver attention modelling~\cite{fridman2016driver}, psychology research~\cite{birmingham2009human}, etc. 

Eye gaze estimation techniques can be broadly classified into two types: \textit{intrusive} and \textit{non-intrusive}. The intrusive technique requires physical contact with user skin or eyes. It includes usage of head-mounted devices, electrodes, or sceleral coils~\cite{xia2007ir,robinson1963method,tsukada2011illumination}. These devices provide accurate gaze estimation but can cause an unpleasant user experience. On the other hand, the non-intrusive technique does not require physical contact~\cite{leo2014unsupervised}. The image processing based gaze estimation methods come under the non-intrusive category. These methods face several challenges, such as occlusion, illumination condition, head pose, specular reflection etc. To overcome these limitations, most of the gaze estimation methods were conducted in constrained environments like fixation of head pose, illumination conditions, camera angle, etc. Moreover, if the method is supervised it require a lot of high-resolution labeled images along with fast and accurate pupil-center localization.

Eye gaze is generally estimated in terms of 2D/3D location or angle in subject's visual space. With the success of supervised deep learning techniques, much progress has been witnessed in most computer vision problems. This is primarily due to the availability of large-sized labeled databases (e.g.: Gaze360~\cite{gaze360_2019}, Eth-X-Gaze~\cite{zhang2020eth}, EVE~\cite{Park2020ECCV} etc.). Furthermore, it has been observed that the labeling of complex vision tasks especially 3D gaze is a noisy and erroneous process. Labelling of 3D gaze dataset requires participant's cooperation and complicated setup. 

Over the past few years, an active research effort is dedicated towards unsupervised, self-supervised and weakly-supervised methods for many real-world applications as it lessen the requirement to acquire the labeled data. Moreover, these methods has recently demonstrated application specific promising results as well~\cite{kolesnikov2019revisiting}. Self-supervised learning techniques are based on a defined \emph{pretext} task which mostly formulated using unlabeled data. In this paper, we define relative pupil location as a pretext task to learn rich representation. The pretext task is mainly inspired by the commonalities between humans' facial features as they shift their gaze from one direction to another. Based on this heuristic, we identify the possible gaze zones. Here, the gaze zones are divided into three regions, i.e, left, right, center. Our pretext task detects the coarse region of interest (aka possible visual attention of the subject) which in turns serves as pseudo labels for self supervised learning. Further, we propose an `Ize-Net' architecture that consists of capsule layer based CNN for learning a discriminating eye-gaze representation. Further, this higher-level semantic understanding is utilized to solve the downstream task. In our case, the downstream tasks include 2D/3D location/angle of eye gaze, visual attention estimation and driver gaze estimation. In brief, we first train our proposed `Ize-Net' model for solving the pretext tasks to learn rich representations which can further be used for solving the downstream tasks of interest. The experimental results show the effectiveness of our technique in predicting the eye gaze as compared to supervised techniques. 

This manuscript \textit{subsumes our earlier work~\cite{dubey2019unsupervised}}. The major changes are as follows: 1) We analyze the effect of learning representation from the eye region only; 2) We add two relevant datasets (MPII and RT-GENE) in the experiment section; 3) We re-evaluate the label through voting and analyze its effect; 4) We adapt our model for driver gaze estimation task (i.e. downstream task); 4) We validate the performance of the `Pretext task' over CAVE dataset.

\begin{figure}[t]
	\centering
	\subfloat{\includegraphics[width = 1in,height=1in]{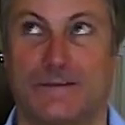}}\hspace{0.01in} 
	\subfloat{\includegraphics[width = 1in,height=1in]{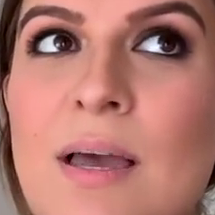}}\hspace{0.01in}
	\subfloat{\includegraphics[width = 1in,height=1in]{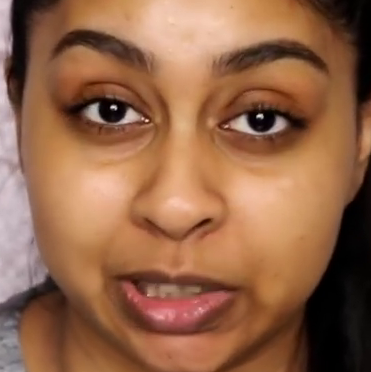}} \\
	\subfloat{\includegraphics[width = 1in,height=1in]{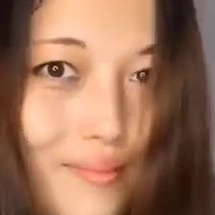}}\hspace{0.01in}
	\subfloat{\includegraphics[width = 1in,height=1in]{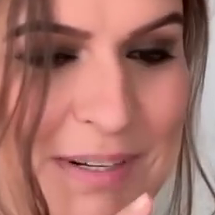}}\hspace{0.01in}
	\subfloat{\includegraphics[width = 1in,height=1in]{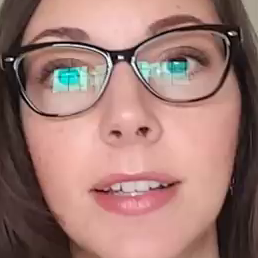}}\hspace{0.01in}
	\\
	\subfloat{\includegraphics[width = 1in,height=1in]{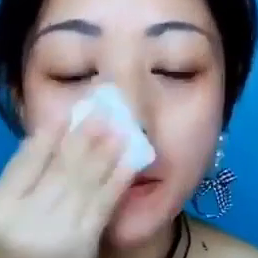}}\hspace{0.01in} 
	\subfloat{\includegraphics[width = 1in,height=1in]{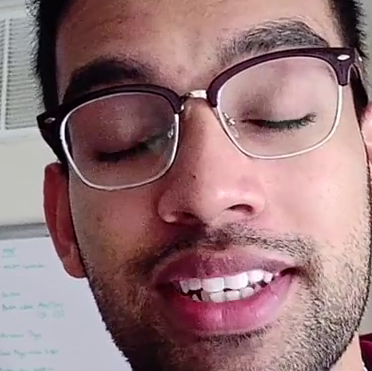}}\hspace{0.01in} 
	\subfloat{\includegraphics[width = 1in,height=1in]{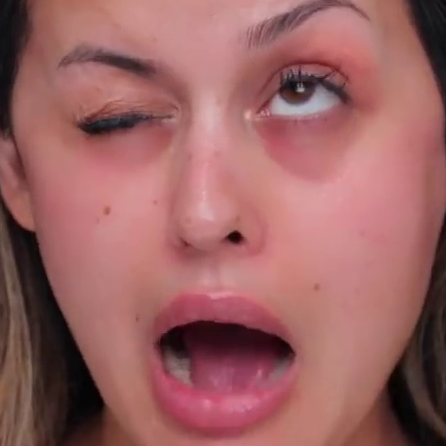}}\hspace{0.01in}
	\caption{Sample images from proposed dataset. Here, we can see that there is huge variation in illumination, facial attributes of subjects, specular reflection, occlusion, etc. First and second rows from top show images for which the gaze region is correctly estimated and third row shows images where gaze region is not correctly estimated. First row; left image subject is looking towards left region. First row; middle image subject is looking towards right region. First row; right image subject is looking towards central region. Second row contains images of challenging scenarios like, occlusion and specular reflection; for which we get correct gaze region estimation. Last row contains images of scenarios where our method fails due to insufficient information for determining correct gaze region. (Image Source: YouTube creative commons)}
	\label{fig:sample_images}
\end{figure}

The \textbf{main contributions} of this paper are as follows:
\begin{itemize}
	\item To the best of our knowledge, we propose \textit{RAZE, a Region guided self supervised gAZE representation learning framework}, one of the first self-supervised technique for eye gaze estimation. The representation learning is guided by a heuristic based auxiliary function i.e. pseudo gaze zone labels.
	
	\item We automatically collect and annotate a dataset (Figure~\ref{fig:sample_images}) of 1,54,251 facial images of 100 different subjects from YouTube videos. The experimental results suggest that this heuristic based annotation method can extract substantial training data for learning robust gaze representation.
	
	\item We propose a capsule layer based deep neural network, `Ize-Net', which is trained on the proposed dataset. The experimental results show that self-supervised techniques can be used for learning rich representation for eye gaze.
	
	\item We demonstrate the effectiveness of learned features for solving downstream tasks as follows: 2D/3D location/angle in subject's visual space, visual attention estimation and driver gaze estimation.
\end{itemize}

The remainder of this paper is organized as follows: Section~\ref{sec:Related Work} describes the relevant prior works. Section~\ref{sec:Proposed Method} presents the details of the proposed pupil-center localization and gaze estimation methods. In Section~\ref{sec:Experiments}, we empirically study the performance of the proposed approach. Section~\ref{sec:Conclusion and Future Work} contains the conclusion, limitation and future work. 

\section{Related Work}
\label{sec:Related Work}
\subsection{Eye Gaze Estimation}
A thorough analysis of gaze estimation literature is mentioned in a recent survey~\cite{ghosh2021Automatic}. Prior works on eye gaze estimation can be broadly classified into hand-crafted and appearance-based methods. We also discuss prior works on pupil center localization as it is relevant to our pretext task.

\subsubsection{Hand-crafted methods} utilize the prior knowledge based on eye anatomy to determine feature values which further help in gaze estimation. Christoph Rasche~\cite{rasche2013curve} propose a labeling functions to identify curved, inflexion and straight segments. With respect to eye gaze, the detection of subject's pupil-centers from simple pertinent features based on shape, geometry, color, and symmetry. These features are then used to extract eye movement information. Morimoto et al.~\cite{morimoto2000pupil} assume a flat cornea surface and proposed a polynomial regression method for gaze estimation. In another interesting work, Zhu et al.~\cite{zhu2002subpixel} extract intensity feature from an image and used a Sobel edge detector to find pupil-center. The gaze direction is further determined via linear mapping function. The main drawback of this method is that the detected gaze direction is sensitive to the head pose; therefore, the users must stabilize their heads. Similarly,  Torricelli et al.~\cite{torricelli2008neural} perform the iris and corner detection to extract the geometric features mapped to the screen coordinates by the general regression neural network. Valenti et al.~\cite{valenti2012accurate,valenti2011combining} estimate the eye gaze by combining the information of eye location and head pose.

\subsubsection{Appearance-based gaze estimation} methods do not explicitly extract the features; instead, these utilize the whole facial/eye image for gaze estimation. Additionally, these methods normally do not require cameras' geometry information and calibration~\cite{lu2017appearance} since the gaze mapping is directly performed on the image content. Fully supervised gaze estimation methods usually require a large number of images to train the estimator. To reduce the training cost, Lu et al.~\cite{lu2014learning} propose a decomposition scheme. It includes the initial gaze estimation and the subsequent compensations for the gaze estimation to perform effectively using training samples. Huang et al.~\cite{huang2015tabletgaze} propose an appearance-based gaze estimation method in which the video captured from the tablet was processed using HoG features and Linear Discriminant Analysis (LDA). Lu et al.~\cite{lu2010novel} propose an eye gaze tracking system which extracted the texture features from the eye regions using the local pattern model. Then a the Support Vector Regressor is utilized to obtain the gaze mapping function. Zhang et al.~\cite{zhang2017mpiigaze} propose GazeNet, which was a deep gaze estimation method. Williams et al.~\cite{williams2006sparse} propose a sparse and semi-supervised Gaussian process model to infer the gaze, which simplifies the process of collecting training data. In brief, the statistical inference based mapping is performed based on K nearest neighbor~\cite{huang2015tabletgaze}, support vector regression~\cite{smith2013gaze}, random forest~\cite{huang2015tabletgaze} and deep learning methods~\cite{krafka2016eye,zhang2017s,zhang2017mpiigaze,jyoti2018automatic,FischerECCV2018,cheng2020gaze,D_2021_CVPR,lrd2022parks}.
	
Several studies~\cite{sugano2014learning,benfold2011unsupervised,zhang2017everyday,santini2017calibme,park2019few,karessli2017gaze,he2019device,lu2015gaze} explore gaze estimation in unsupervised and semi-supervised settings to reduce the burden of data annotation. These approaches are mainly based on `learning-by-synthesis'~\cite{sugano2014learning}, hierarchical generative models~\cite{wang2018hierarchical}, conditional random field~\cite{benfold2011unsupervised}, unsupervised gaze target discovery~\cite{zhang2017everyday}, gaze redirection~\cite{yu2019improving}, multi-task learning/MTGLS~\cite{ghosh2021mtgls}, weakly supervised using via `Looking At Each Other (LAEO)'~\cite{kothari2021weakly}, cross-modal supervision~\cite{ghosh2022av} and few-shot learning~\cite{park2019few}. MTGLS~\cite{ghosh2021mtgls} framework leverages complementary signals via the line of sight of the pupil, the head-pose and the eye dexterity.

In literature, the domain specific knowledge is also leveraged to get strong complimentary information. These information includes facial landmark~\cite{yu2018deep}, screen saliency~\cite{Park2020ECCV,wang2019inferring}, depth~\cite{lian2019rgbd}, headpose~\cite{zhu2017monocular}, segmentation mask~\cite{wu2019eyenet} and uncertainty~\cite{gaze360_2019}. Unlike this, our study focuses on automatic gaze region labeling as pretext task to reduce the annotation burden as well as infer coarse to fine gaze adaptation.

\subsubsection{Pupil Center Localization}
Prior works on pupil-center localization can be broadly classified into two categories based on active and passive techniques~\cite{leo2014unsupervised}. The active pupil-center localization methods utilize dedicated devices to locate the pupil-center by infrared camera~\cite{xia2007ir}, contact lenses~\cite{robinson1963method} and  head-mounted devices~\cite{tsukada2011illumination}. These devices require a pre-calibration phase to perform accurately. These are generally very expensive and cause an uncomfortable user experience. The passive eye localization methods try to gather information from the supplied image/video-frame regarding the pupil-center. Valenti et al.~\cite{valenti2012accurate} have used identical images to infer circular patterns and used machine learning for the prediction task. An open eye can be peculiarly defined by its shape and its components like iris and pupil contours. The structure of an open eye can be used to localize it in an image. Such methods can be broadly divided into voting-based methods~\cite{kim1999vision,perez2003precise} and model fitting methods~\cite{daugman2003importance,hansen2005eye}. Although these methods seem very intuitive, but it fails to provide good accuracy in real world secnarios. Several machine learning based pupil-center localization methods have also been proposed. One such method was proposed by Campadelli et al.~\cite{campadelli2009precise}, in which they used two Support Vector Machines (SVM) and trained them on properly selected Haar wavelet coefficients. Markuvs et al.~\cite{markuvs2014eye} use randomized regression trees for pupil localization. Prior works on pupil-center localization is mainly based on geometric feature which gives accurate results for images captured under an controlled environment. The geometric models are mainly based on physical measurements; it generalizes quite easily to new subjects with very few prior annotated data.
	
\subsection{Self-supervised Learning Paradigm}
Self-supervised learning attracts many researchers for its superior performance gain on different vision based emerging topics in the past few years~\cite{zhao2022unsupervised}. Self-supervised representation learning mainly leverages input data itself for supervision and infers for any relevant downstream tasks. One recent study~\cite{goyal2019scaling} shows that by leveraging various attributes of the data (for example: input data size), self-supervised technique can largely match or even exceed the performance of supervised pre-training on a variety of tasks such as object detection, surface normal estimation (3D) etc. Kocabas et al.~\cite{kocabas2019self} show that even without any 3D ground truth data and the knowledge of camera extrinsics, multi view images can be leveraged to obtain self supervision. Definition of appropriate pretext task is very crucial for self-supervised learning. Misra et al.~\cite{misra2020self} develop pretext-invariant representation learning that learns invariant representations based on pretext tasks. A recent survey~\cite{jing2020self} on self-supervised approach depicts the potential to explore this domain. 

In gaze representation learning domain, Yu et al.~\cite{yu2019unsupervised} uses subject specific gaze redirection as a pretext task to learn strong representation. Swapping Affine Transformations (SwAT)~\cite{farkhondeh2022towards} is the extended version of Swapping Assignments Between Views (SwAV), a popular self supervised learning framework. It is used for gaze representation learning using different augmentation techniques. Following this trend, our approach also defines a pretext task of gaze region classification based on relative pupil location to learn efficient representation for eye gaze estimation.

\begin{table*}[t]
	\centering
	\caption{A statistical overview of gaze datasets in literature.}
	\label{subject_comparison}
	\scalebox{1.0}{
		\begin{tabular}{c||c|c|c|c|c|c|c|c|c|c|c}
		\toprule[0.4mm]
		\rowcolor{mygray}
			\textbf{Datasets}  & \textbf{\begin{tabular}[c]{@{}c@{}}Gi4E\\~\cite{villanueva2013hybrid}\end{tabular}} & \textbf{\begin{tabular}[c]{@{}c@{}}RT-GENE\\~\cite{FischerECCV2018}\end{tabular}} & \textbf{\begin{tabular}[c]{@{}c@{}}CAVE\\~\cite{smith2013gaze}\end{tabular}} & \textbf{\begin{tabular}[c]{@{}c@{}}OMEG\\~\cite{he2015omeg}\end{tabular}} &\textbf{\begin{tabular}[c]{@{}c@{}}MPIIGaze\\~\cite{zhang15_cvpr}\end{tabular}}&\textbf{\begin{tabular}[c]{@{}c@{}}TabletGaze\\~\cite{huang2015tabletgaze}\end{tabular}}&\textbf{\begin{tabular}[c]{@{}c@{}}GazeCapture\\~\cite{cvpr2016_gazecapture}\end{tabular}}&\begin{tabular}[c]{@{}c@{}}\textbf{Gaze 360}\\~\cite{gaze360_2019} \end{tabular}& \textbf{\begin{tabular}[c]{@{}c@{}}ETHX-Gaze \\ \cite{zhang2020eth} \end{tabular}} & \textbf{\begin{tabular}[c]{@{}c@{}}EVE\\ \cite{Park2020ECCV} \end{tabular}} &\textbf{\begin{tabular}[c]{@{}c@{}}RAZE \end{tabular}} \\ \hline \hline
			Subjects                                                                          & 103                                                         & 15                                                                        & 56                                                                                                       & 50  & 15 & 41 & 1450               & 238 & 110 & 54 & 100           \\ 
			\begin{tabular}[c]{@{}c@{}} Total  Images\end{tabular}                                                                              & 1K                                                         & 122K                                                                        & 5K                             & 44K             & 213K  & 100K  &   2445K             & 172K & 1083K & 12308K & 154K           \\ 
	\bottomrule[0.4mm]		
	\end{tabular}}
	
\end{table*}

\section{Method}
\label{sec:Proposed Method}
In this section, we describe the overview of the proposed self-supervised gaze region estimation method. Accurate gaze direction estimation usually depends on several factors such as exact locations of the pupil centers, head-pose, eye blink and subject specific appearance. However, the existing benchmark datasets are curated in constrained environments. Thus, instead of limiting ourselves to these data, we web-crawled YouTube videos having creative common licence. Our proposed frame-work,  RAZE is guided by pseudo-gaze zone classification objective which can further be adapted to other downstream tasks. Figure~\ref{fig:pipeline} refects the overview of the proposed framework.

\subsection{Representation Learning Framework}
\noindent \textbf{Preliminaries.}
Given a detected face $\mathbf{x}$ from dataset $\mathcal{D}$, we localize the pupil-centers (i.e. $(p_{x}^{l},p_{y}^{l})$ and $(p_{x}^{r},p_{y}^{r})$) of the concerned subject at first. Further, the relative position of the pupils are utilized as a pretext task to estimate the eye gaze region $\mathbf{e} \in \mathbb{R}^3$ (i.e. \emph{left}, \emph{right} and \emph{center}) of the subject. The RAZE framework learn the meaningful representation of the eye region via `Ize-Net' network parameterised by $\mathcal{F}_{\phi}$ . $\mathcal{F}_{\phi}$ maps the input $\mathbf{x}$ to feature space $\mathbf{z}$ by $\mathcal{F}_{\phi}:\mathbf{x} \to \mathbf{z}$, where $ \mathbf{z}\in \mathbb{R}^d $. Later, the latent representation is mapped to the label space by $\mathcal{F}_{\theta}:\mathbf{z} \to \mathbf{e}$. The workflow of the whole self-supervised paradigm is summarized in Algorithm~\ref{alg:RAZE}. The rest of the section contains details of each stages mentioned in Algorithm~\ref{alg:RAZE}.  

\noindent \textbf{Pupil-Center Localization.}
The first stage of our proposed method is pupil center localization. Accurate pupil-center localization plays an important role in eye gaze estimation. We take face image as input and extract eye-regions from this image, using the facial landmarks obtained by the Dlib-ml library~\cite{king2009dlib}. Further processing is performed on the extracted eye images. We localize the pupil-center using a three stage method, i.e., blob center detection~\cite{lin2011pupil} and CHT~\cite{daway2018pupil}, and take the average of the pupil-centers obtained by both of the methods to calculate the final pupil-center. The steps of the proposed pupil-center localization method are as follows (See Algorithm~\ref{alg:pupil_loc}):
\begin{enumerate}
	\item Extract eyes using facial landmark information.
	\item Apply OTSU thresholding on the extracted eyes to take the advantage of unique contrast property of eye region while pupil circle detection.
	\item Apply the method of blob center detection on extracted iris contours to calculate 'primary' pupil-centers.
	\item Crop regions near these centers, to perform the center rectification task. The crop length is decided by applying equation (1). 
	\begin{equation}
	\texttt{Crop}\ \texttt{len.} = \dfrac{\texttt{Height}\ \texttt{of}\ \texttt{eye}\ \texttt{contour}}{\texttt{2}} + \texttt{offset}
	\end{equation}
	\item Compute Adaptive thresholding and apply Canny edge detector~\cite{canny1986computational} to make the iris region more prominent.
	\item Apply CHT over the edged image to find secondary pupil-centers.
	\item Compute average of primary and secondary pupil-centers to finalize the value for pupil-centers.
\end{enumerate}
The detected pupil centers are utilized for the pretext task which is described next. 

\begin{algorithm}[tb]
\caption{Training Procedure for RAZE}
\label{alg:RAZE}
\begin{algorithmic}[1]
\Require{$\mathcal{F}_\phi$, $\mathcal{F}_{\theta}$, and $\mathcal{D}$}
\For{\texttt{$\mathbf{n}$ epochs}} \Comment{RAZE Training}
\State $\mathbf{e} \gets \texttt{Heuristic }(\mathbf{x})$ \Comment{Pretext Task}
\State $\mathbf{z}\gets \mathcal{F}_{\phi}(\mathbf{x})$
\State $\mathbf{e'}\gets \mathcal{F}_{\theta}(\mathbf{z})$
\State $\mathcal{L}_0 = \mathcal{L}_{\texttt{gaze-region}} (\mathbf{e}, \mathbf{e'})$ 
\State $\{\phi, \theta\} \gets \triangledown_{\{\phi, \theta\}} \mathcal{L}_0$
\EndFor
\For{\texttt{$\mathbf{n}$ epochs}} \Comment{Downstream Adaptation}
\State $\mathbf{z} \gets \mathcal{F}_{\phi}(\mathbf{x'})$\Comment{$x' \in \mathcal{D}$}
\State $\mathbf{y'} \gets \mathcal{F}_{\theta} (\mathbf{z})$ 
\State $\mathcal{L}_1 = \mathcal{L}_{\texttt{FT/LP}} (\mathbf{y}, \mathbf{y'})$  \Comment{Dataset specific Fine-Tuning or Linear Probing}
\State $\{\phi$ ,/or $\theta\}\gets \triangledown_{\{\phi,/or\theta\}} \mathcal{L}_1$ 
\EndFor
\end{algorithmic}
\end{algorithm}

\begin{figure*}[t]
	\centering
	\includegraphics[height=5cm,width=\linewidth]{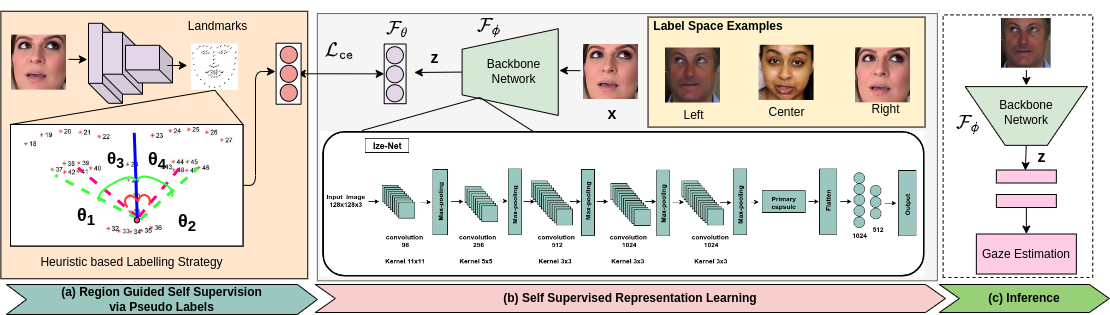}
	\caption{Overview of the proposed pipeline. From left to right, we show \textit{(a) Region Guided Self Supervision via Pseudo Labels:} The proposed RAZE module first perform pseudo labelling of the detected faces based on facial landmarks. Angles $\theta_1$ and $\theta_2$ are used to estimate eye gaze region and angles $\theta_3$ and $\theta_4$ are used for head pose estimation. (Refer Sec.~\ref{sec:Proposed Method}-A Pretext Task for more details); \textit{(b) Self Supervised Representation Learning:} RAZE framework consists of the backbone network aka `Ize-Net' which maps input image to the label space.Few label space examples are also shown in yellow bounding box (Refer Sec.~\ref{sec:Proposed Method}-A for more details); \textit{(c) Inference:} We use Linear Probing (LP), Fine-Tuning (FT) for adapting to different datasets and tasks. } 
	\label{fig:pipeline}
\end{figure*}

\noindent \textbf{Pretext Task: Heuristic for Eye Gaze Region Estimation.}
Pretext task is the second step of our proposed self supervised paradigm. The relative position of the pupil-centers is the most decisive feature of the face to determine gaze direction. Eye, head movement and their relative motion determines the direction of the `coarse-level' eye gaze. Thus, by using the relative position of both the pupil-centers, we can determine the possible regions where the subject is looking. When a subject looks towards his/her left, both the eyes' iris shift towards left. To utilize this unique characteristic, we compare the angles formed when we join the left pupil-center with the nose and nose with vertical; with the angle formed when we join the right pupil-center with the nose and nose with vertical. These angles are demonstrated in Figure~\ref{fig:pipeline} as angles $\theta_1$ and $\theta_2$. For a subject to look towards his/her left region, the left eye angle $\theta_1$ has to be bigger than the right eye angle $\theta_2$. This intuitive heuristic is used to detect the coarse-level gaze region (left, right, or center) in which the subject is looking. Empirically, the proposed method is immune to head movements within the range of \ang{-10} to \ang{10}. 

The eye corners remain fixed with the eye movement. We utilize the eye corner points given by the Dlib-ml library to determine the head pose direction, in the same way as we determine the eye gaze region. The angles used to determine the head pose direction are demonstrated in Figure~\ref{fig:pipeline} as angles $\theta_3$ and $\theta_4$. For example, when the subject's head pose is left the $\theta_4$ is greater than $\theta_3$. By using this pretext task, we collect and annotate a large scale YouTube data described later.

\noindent \textbf{Overall RAZE Loss.}
Algorithm~\ref{alg:RAZE} describes the training procedure of the proposed RAZE. The overall learning is guided by the following objective functions: $\mathcal{L}_{\texttt{gaze-region}}= \mathcal{L}_{\texttt{{ce}}}$. 

\noindent Here,  $\mathcal{L}_{\texttt{{ce}}}$ is the standard cross entropy loss for three gaze zone/regions.

\subsection{Evaluation Protocol for Self-Supervision}

Following the standard evaluation protocols for self supervised learning paradigms, we also adopt Linear Probing (LP)~\cite{zhang2016colorful,he2020momentum,caron2021emerging} and Fine-Tuning (FT) for downstream adaptation~\cite{caron2021emerging}. For LP, we incorporate data augmentation strategy in terms of random resize, random crops and flipping horizontally during training phase. In gaze estimation, the ground truth gaze labels change its sign while performing horizontal flipping operation. While adapting via LP the weights $\phi$ is frozen and only the label space parameters i.e. $\mathcal{F}_{\theta}$ are updated. The downstream adaptation process is enforced by the appropriate loss function for different tasks. To be more specific, for 3D gaze estimation the following loss is incorporated $\mathcal{L}_{\texttt{3D\ gaze}} = \frac{\mathbf{g}}{||\mathbf{g}||_2}. \frac{\mathbf{g'}}{||\mathbf{g'}||_2}$ where, $\mathbf{g}$ and $\mathbf{g'}$ are ground truth and predicted labels.

For FT, instead of $\mathcal{F}_{\theta}$, all of the parameters of RAZE are updated. However, the training is started with the pretext tasked based pre-trained weights.

\begin{algorithm}[tb]
\caption{Pupil Center Localization}
\label{alg:pupil_loc}
\begin{algorithmic}[1]
\For{\texttt{$\mathbf{n}$ images}} \Comment{Pupil Center Localization}
    \State $\mathbf{Eyes} \gets \texttt{Dlib-ml}\ (\mathbf{x})$ \Comment{ Eye localization via Facial landmarks} 
    \State $\mathbf{Iris}\gets  \texttt{OTSU}\ (\mathbf{Eyes})$ \Comment{OTSU Method} 
    \State $\mathbf{P_{p}}\gets  \texttt{Blob\ Center\ Detection}\ (\mathbf{Iris})$  \Comment{`Primary' Pupil-Center}
    \State  $\texttt{Crop-Length} = \dfrac{\texttt{Height}\ \texttt{of}\ \texttt{eye}\ \texttt{contour}}{\texttt{2}} + \texttt{offset}$
    \State ROI $\gets$ Crop regions near Pupil-Center
    \State Adaptive Thresholding (ROI) \Comment{Iris Center
Rectification} 
    \State  $\mathbf{P_{s}}\gets$ CHT (Canny Edge (ROI)) \Comment{`Secondary' Pupil-Centers} 
    \State $\mathbf{\mathbf{P_{c}}}\ =\ \dfrac{\mathbf{P_{p}} + \mathbf{P_{s}}}{2}$
\EndFor 
\end{algorithmic}
\end{algorithm}

\section{Experimental Protocols}
\label{sec:Experiments}
For all of our experiments, we use the Keras deep learning library with the Tensorflow backend. The proposed deep model for eye gaze estimation was trained and tested on Titan Xp GPU.

\begin{table}[t]
	\centering
	\caption{The categorical distribution of the proposed dataset.}
	\label{data_stat}
	\scalebox{1.0}{
		\begin{tabular}{l||c|c|c|c}
			\toprule[0.4mm]
			\rowcolor{mygray}
			\textbf{RAZE Dataset} & \textbf{Center} & \textbf{Left} & \textbf{Right} & \textbf{Total} \\ \hline \hline
			Train set        & 32,450           & 38,230         & 37,338          & 108,018         \\ 
			Validation set   & 14,008           & 16,584         & 15,641          & 46,233          \\ 
			Total            & 46,458           & 54,814         & 52,979          & 1,54,251         \\ 
			\bottomrule[0.4mm]
	\end{tabular}}
\end{table}

\noindent \textbf{Benchmark Datasets.} We evaluate the proposed method RAZE on five benchmark datasets: \textbf{CAVE}~\cite{smith2013gaze}, \textbf{MPII}~\cite{zhang2017mpiigaze}, \textbf{TabletGaze}~\cite{huang2015tabletgaze}, \textbf{RT-GENE}~\cite{fischer2018rt} and \textbf{DGW}~\cite{ghosh2021speak2label}. CAVE~\cite{smith2013gaze} dataset has 5,880 high resolution images of 56 subjects. The dataset is collected in a constrained lab environment. The data is labelled for 21 different gaze directions and head-poses for each subject. MPII~\cite{zhang2017mpiigaze} dataset is collected from 15 subjects performing everyday activity before a laptop. The dataset contains 213,659 images collected over a three-month window. TabletGaze~\cite{huang2015tabletgaze} is relatively unconstrained dataset of 51 subjects. The gaze direction is mapped with 4 different postures and 35 gaze locations. This dataset is also collected in an indoor environment. Similarly, RT-GENE dataset~\cite{fischer2018rt} is also recorded in a naturalistic environment. The ground truth annotation is assigned using a motion capture system connected with eye-tracking glasses. DGW~\cite{ghosh2021speak2label} is a large scale driver gaze zone estimation dataset. DGW contains data from 338 subjects fixating their gaze `inside a car' scenario with variation in illumination, occlusion etc. We validate the proposed pupil-center localization method (See Algorithm~\ref{alg:pupil_loc}) on BioID dataset~\cite{jesorsky2001robust}. BioID is a publicly available dataset which contains 1,521 frontal face images of 23 subjects.

\noindent \textbf{Automatic Dataset Collection Paradigm.}
In recent years, several gaze estimation datasets have been proposed~\cite{gaze360_2019,FischerECCV2018}. Most of the datasets are collected in more or less restricted environment. Moreover, few of these datasets may contain very little of images in terms of head poses, illumination, number of images, collection duration per subject and camera quality. To demonstrate the adaptability of our proposed self supervised method, we collect a dataset containing 154,251 facial images belonging to 100 different subjects from YouTube (having creative common license). The overall statistic of our dataset is shown in Table~\ref{data_stat}. We download different types of videos from YouTube. These videos belong to different categories, where a single (or multiple) subject(s) is seen on the screen at a time, like news reporting, makeup tutorials, speech videos, doing meditation etc. We have considered every third frame of the collected videos for dataset creation. The dataset has been split into training and validation sets with 70\% and 30\% uniform partitions over the subjects for the training purpose. The overview of our proposed dataset is shown in Figure~\ref{fig:sample_images}. In this figure, we can observe that our dataset contains a huge variety of images with varying illumination, occlusion, blurriness, color intensity, etc. Table~\ref{subject_comparison} provides the comparison of the state-of-the-art gaze datasets with our proposed dataset. \textit{Please note that the dataset is available upon request.}

\noindent \textbf{Implementation Details.}
\textit{1. Network Architecture:} The architecture of the proposed `Ize-Net' network is shown in Figure~\ref{fig:pipeline}. The network uses a primary capsule component combined with a series of convolution layers. The motivation of using capsule block stems from the superior performance of capsule networks~\cite{sabour2017dynamic} in handling relative location of an object's parts. Our network is trained using images of size $128\times128\times3$. We take the entire face as input instead of only the eye region. According to~\cite{zhang2017s}, gaze can be more accurately predicted when the entire face is considered. Our proposed network contains five convolution layers. Each convolution is followed by batch normalization and max-pooling. For batch normalization, we use 'ReLU' as the activation function. For max-pooling kernel of size ($2\times2$) was used. The stride of ($1\times1$) is considered for each layer. After the convolution layers, we append primary capsule, whose job is to take the features learned by convolution layers and produce combinations of the features to consider face symmetry into account. The primary capsule output is flattened and fed to fully-connected layers of dimension 1024 and 512. In the end, we apply softmax activation to produce the final output which is gaze regions (i.e. left, right and center). 

\noindent\textit{2. Linear Probing(LP) and Fine Tuning(FT) details:} To linear probe the base model for prospective datasets, we add two Fully-Connected (FC) layers (dimension 256) at the end of the proposed Ize-Net network. For LP, we  demonstrate the impact of weight freezing (at different level) on gaze estimation performance. The last 8 layers, last 12 layers, and complete network are fine-tuned in succession for the empirical analysis of results. For fine-tuning the network on the Tablet Gaze dataset, we used a learning rate of 0.0001 with 10 epochs, and for the other datasets, we used a learning rate of 0.0001 with 15 epochs. During fine-tuning, the mean square error loss function as well as cosine similarity is implemented following the respective evaluation protocols mentioned in prior literature. We fine-tune the Ize-Net on the DGW dataset using the SGD optimizer for 20 epochs with a learning rate of 0.0001, the decay of $1 \times e ^{-6}$ per epoch and momentum of 0.9. 

We additionally evaluate a weighted nearest neighbour classifier (k-NN)~\cite{caron2021emerging} on the DGW data. The weights of the Ize-Net is frozen and the penultimate layer's feature is extracted for training. The k-NN classifier uses similarity matching operation along with voting strategy in the latent space to get the predicted label. Empirically, this analysis works for $\sim$ 13-15 NN over several iterations. 

\noindent \textbf{Evaluation Metrics.}
For quantitative evaluation of the gaze region estimation, we use class-wise accuracy (in \%). Following each database's evaluation protocol, we follow `leave-one-person-out' for MPII, cross-validation for CAVE and TabletGaze; and 3-fold evaluation for RT-GENE dataset. Additionally, we compute angular error (in \textdegree) except for the TabletGaze dataset, for which we compute the error in cm (similar to ~\cite{huang2015tabletgaze}). To compare with the state-of-the-art methods, we use similar evaluation protocols mentioned in those studies.

\begin{table*}[t]
	\centering
	\caption{\textbf{Results on Tablet Gaze (in cm)} with comparison to baselines~\cite{smith2013gaze}. Effectiveness of learnt features in Ize-Net (Pre-trained on the collected data) is demonstrated by the fine tuning the network and by training a SVR over various FC layer features. * methods are supervised.}
	\label{tg_compare}
	\scalebox{0.9}{
		\begin{tabular}{c|c|c|c|c|c|c|c|c|c|c|c|c}
			\toprule[0.4mm]
			\rowcolor{mygray}
			\begin{tabular}[c]{@{}c@{}} \textbf{Methods} \end{tabular}& \begin{tabular}[c]{@{}c@{}}\textbf{Raw pixels*}\\~\cite{huang2015tabletgaze}\end{tabular} & \begin{tabular}[c]{@{}c@{}}\textbf{LoG*}\\~\cite{huang2015tabletgaze} \end{tabular} & \begin{tabular}[c]{@{}c@{}}\textbf{LBP*}\\~\cite{huang2015tabletgaze} \end{tabular} & \begin{tabular}[c]{@{}c@{}}\textbf{HoG*}\\~\cite{huang2015tabletgaze} \end{tabular} & \begin{tabular}[c]{@{}c@{}}\textbf{mHoG*}\\~\cite{huang2015tabletgaze} \end{tabular} & \begin{tabular}[c]{@{}c@{}} \textbf{\cite{jyoti2018automatic}}* \end{tabular}& \begin{tabular}[c]{@{}c@{}}\textbf{RAZE}\\ \textbf{(Full Network}\\ \textbf{ Fine Tuning})\end{tabular} & \begin{tabular}[c]{@{}c@{}}\textbf{RAZE} \\ \textbf{(last} \\ \textbf{12 layers} \\\textbf{fine-tuning)} \end{tabular} & \begin{tabular}[c]{@{}c@{}}\textbf{RAZE} \\ \textbf{(last} \\ \textbf{8 layers} \\\textbf{fine-tuning)}\end{tabular} & \begin{tabular}[c]{@{}c@{}}\textbf{RAZE} \\ \textbf{(last} \\ \textbf{8 layers} \\\textbf{fine-tuning)}\\ \textbf{with eye patch}\end{tabular} &  \begin{tabular}[c]{@{}c@{}}\textbf{RAZE} \\ \textbf{Layer (34)} \\ \textbf{+ SVR} \end{tabular} & \begin{tabular}[c]{@{}c@{}}\textbf{RAZE} \\ \textbf{Layer (31)} \\ \textbf{+ SVR} \end{tabular} \\ \hline \hline
			k-NN               & 9.26       & 6.45 & 6.29 & 3.73 & 3.69 & \multirow{4}{*}{2.61}& \multirow{4}{*}{2.36}                                                                     & \multirow{4}{*}{3.31}                                                                                    & \multirow{4}{*}{3.26}                                                                 & 
			\multirow{4}{*}{2.80}
			& \multirow{4}{*}{2.42}          
			& \multirow{4}{*}{2.48}                                           \\ \cline{1-6}
			RF                 & 7.2        & 4.76 & 4.99 & 3.29 & 3.17 &                                                                                            &                                                                                                          &                                                                                                &            &            &                                                        &                                                                   \\ \cline{1-6}
			GPR                & 7.38       & 6.04 & 5.83 & 4.07 & 4.11 &                                                                                            &                                                                                                          &          &                                                                                                  &           &                                                        &                                                                   \\ \cline{1-6}
			SVR                & -          & -    & -    & -    & 4.07 &                                                                                            &                                                                                                          &                                                        &                                                    &         &                                                          &                                                                   \\ 
			\bottomrule[0.4mm]
	\end{tabular}}
\end{table*}
	
\begin{table*}[!htbp]
	\centering
	\caption{\textbf{Results on the CAVE dataset} (Pre-trained on the collected data) using the angular deviation, calculated as $ \texttt{mean}\ \texttt{error}\ $(in \textdegree)$\pm\  \texttt{standard\ deviation} $ (in \textdegree). It is interesting to note that the eye patch region based learnt representation performs best. * methods are supervised.}
	\label{cave_compare}
	\scalebox{1}{
		\begin{tabular}{c|c|c|c|c|c}
			\toprule[0.4mm]
			\rowcolor{mygray}
			\textbf{Calibration}                                                                          & \multirow{1}{*}{\textbf{Method}} & \multicolumn{2}{c|}{\textbf{$\ang{0}$ yaw angle}} & \multicolumn{2}{c}{\textbf{Full Dataset}} \\ \cline{1-1} \cline{3-6} 
			\multirow{4}{*}{\begin{tabular}[c]{@{}c@{}}5 point system\\ (cross arrangement)\end{tabular}} &                                 \cellcolor{mygray} & \cellcolor{mygray}\textbf{X}           &\cellcolor{mygray} \textbf{Y}          & \cellcolor{mygray}\textbf{X}           & \cellcolor{mygray} \textbf{Y}          \\ \cline{2-6} 
			& Skodras et al.~\cite{skodras2015visual}*           & $ 2.65 \pm 3.96 $          &  $ 4.02 \pm 5.82  $         & N/A                  & N/A                 \\ \cline{2-6} 
			& Jyoti et al.~\cite{jyoti2018automatic}*                        &  $ 2.03 \pm 3.01  $          &  $ 3.47 \pm 3.99       $    & N/A                  & N/A                 \\ \cline{2-6} 
			& \textbf{RAZE} (full face)                            &  $ 2.94\pm 2.16      $      &  $ 2.74\pm1.92  $           &  $ 1.67\pm1.19  $            &  $ 1.74\pm1.57    $         \\ \cline{2-6} 
			& \textbf{RAZE} (eye patch)                            &  $ 2.65 \pm 1.70  $          &  $ 2.16 \pm 1.44    $         &  $  0.98 \pm 0.74  $            &  $ 1.05 \pm 0.73   $          \\ 
			\bottomrule[0.4mm]
	\end{tabular}}
\end{table*}

\begin{table}[t]
	\centering
	\caption{\textbf{Results on RT-GENE dataset~\cite{fischer2018rt} (in \textdegree)} which is pre-trained on the RAZE data. * methods are supervised.}
	\label{tab:rt-gene}
	\begin{tabular}{c|c|c|c|c}
		\toprule[0.4mm]
		\rowcolor{mygray}
		\textbf{\begin{tabular}[c]{@{}c@{}}Single Eye\\~\cite{zhang15_cvpr}* \end{tabular}} & \textbf{\begin{tabular}[c]{@{}c@{}}Spatial \\ weights \\ CNN\\~\cite{zhang2017s}*\end{tabular}} & \textbf{\begin{tabular}[c]{@{}c@{}}Spatial \\ weights\\ CNN \\ (ensemble)~\cite{fischer2018rt}*\end{tabular}} & \textbf{\begin{tabular}[c]{@{}c@{}}4 model \\ ensemble~\cite{fischer2018rt}*\end{tabular}} & \textbf{RAZE}  \\ \hline \hline
		13.4                & 8.7                                                                     & 8.7                                                                                   & 7.7                                                                  & 6.1                 \\ 
		\bottomrule[0.4mm]
	\end{tabular}
\end{table}

\begin{table}[t]
	\centering
	\caption{\textbf{Results on MPII dataset~\cite{fischer2018rt} (in \textdegree)} which is pre-trained on the RAZE data. * methods are supervised.}
	\label{tab:mpii}
	\scalebox{1.0}{
		\begin{tabular}{c|c|c|c|c|c|c}
			\toprule[0.4mm]
		\rowcolor{mygray}
			\textbf{\begin{tabular}[c]{@{}c@{}}Single\\ Eye\\~\cite{zhang15_cvpr}*\end{tabular}} & \textbf{\begin{tabular}[c]{@{}c@{}}iTracker \\~\cite{krafka2016eye}*\end{tabular}} & \textbf{\begin{tabular}[c]{@{}c@{}}Two\\ Eyes \\~\cite{fischer2018rt}*\end{tabular} } & \textbf{\begin{tabular}[c]{@{}c@{}}iTracker \\ (AlexNet)\\~\cite{krafka2016eye}*\end{tabular}} & \textbf{\begin{tabular}[c]{@{}c@{}}Single\\ Face \\~\cite{fischer2018rt}*\end{tabular}} & \textbf{\begin{tabular}[c]{@{}c@{}}Spatial \\ weights\\ CNN \\~\cite{zhang2017s}*\end{tabular}} & \textbf{RAZE} \\ \hline \hline
			6.7                 & 6.2               & 6.2               & 5.6                                                                    & 5.5                  & 4.8                                                                     & 5.0           \\ 
			\bottomrule[0.4mm]
	\end{tabular}}
	
\end{table}

\begin{table}[t]
	\centering
	\caption{Fine-tuning result on DGW dataset~\cite{ghosh2021speak2label} for driver gaze estimation. * methods are supervised.}
	\label{tab:speak2label}
	\begin{tabular}{l||c|c}
		\toprule[0.4mm]
		\rowcolor{mygray}
		\textbf{Method}   & \textbf{Val. Accuracy} & \textbf{Test Accuracy} \\ \hline \hline
Vasli et al.~\cite{vasli2016driver}*     &   {52.60}                                                                        &   {50.41}                                                                   \\ 
Tawari et al.~\cite{tawari2014driver}*    &   {51.30}                                                                        &   {50.90}                                                                   \\ 
Fridman et al.~\cite{fridman2015driver}*     &   {53.10}                                                                        &   {52.87}                                                                   \\ 
\begin{tabular}[c]{@{}c@{}}Vora et al.~\cite{vora2017generalizing} (Alexnet face)* \end{tabular} &  {56.25  }                                                                         &  {57.98 }                                                                   \\ 
\begin{tabular}[c]{@{}c@{}}Vora et al.~\cite{vora2017generalizing} (VGG face)*\end{tabular}     &  {58.67}                                                                        &  {58.90 }                                                                   \\ 
SqueezeNet~\cite{iandola2016squeezenet}*                                                                             &  {59.53}                                                                           &  {59.18 }                                                                    \\ 
Ghosh et al.~\cite{ghosh2021speak2label}*                                                                              &  {60.10}                                                                           &  {60.98}                                                                    \\ 
Inception V3~\cite{DBLP:journals/corr/SzegedyVISW15}*                                                                           &  {67.93}                                                                           &  {68.04}                                                                    \\ 
Vora et al.~\cite{vora2018driver}*                                                         &  { 67.31}                                                       &  {68.12}                                                \\ 
ResNet-152~\cite{DBLP:journals/corr/HeZRS15}*                                                                              &  {68.94}                                                                           &  {69.01 }                                                                   \\ 
\begin{tabular}[c]{@{}c@{}}Yoon et al.~\cite{yoon2019driver} (Face + Eyes)* \end{tabular}     &   {70.94}                                                                        &   {71.20}                                                                   \\ 

\begin{tabular}[c]{@{}c@{}}Stappen et al.~\cite{stappen2020x}*\\\end{tabular}     &   {71.03}                                                                        &   {71.28}                                                                   \\
\begin{tabular}[c]{@{}c@{}}Lyu et al.~\cite{lyu2020extract}*\\\end{tabular}     &   {85.40}                                                                        &   {81.51}                                                                   \\ 
\begin{tabular}[c]{@{}c@{}}Yu et al.~\cite{yu2020multi}* \\\end{tabular}     &   {80.29}                                                                        &   {82.52}                                                                   \\
RAZE (k-NN)     & 62.50             & 63.82       \\ 
RAZE (LP)     & 72.10             & 73.02       \\ 
RAZE (FT)     & 80.50             & 81.82       \\ 
	\bottomrule[0.4mm]
	\end{tabular}
\end{table}

\section{Results}
We conduct comprehensive quantitative and qualitative analysis to validate our method on five publicly available benchmark datasets. We have also performed extensive ablation studies to show the impact of different components of the proposed pipeline.

\subsection{Downstream Task Specific Adaptation}
The `Ize-Net' network is trained on the proposed dataset for the task of gaze region estimation. We adapt the proposed method on 3D gaze estimation and driver gaze zone estimation tasks described below.

\noindent \textbf{`Coarse-to-fine' gaze estimation:} The learned data representation is linear-probed (LP) and fine-tuned (FT) on four benchmark gaze estimation datasets (i.e. TabletGaze~$\rightarrow$ Table~\ref{tg_compare}, CAVE~$\rightarrow$ Table~\ref{cave_compare}, MPII~$\rightarrow$ Table~\ref{tab:mpii} and RT-GENE~$\rightarrow$ Table~\ref{tab:rt-gene}) for determining the exact gaze location. Here, gaze location indicates the 3D/2D location/gaze-angle of the concerned subject. 

In TABLE~\ref{tg_compare}, we incorporate the weight freezing strategy at different levels to determine the optimal layer for rich feature extraction. The last 8 layers, last 12 layers, and complete network are fine-tuned in succession for the empirical analysis of results. The empirical analysis suggest that the full network fine-tuning performs best for downstream adaptation. Even it outperforms supervised state-of-the-art~\cite{jyoti2018automatic} significantly ($2.61$cm $\rightarrow$ $2.36$ cm, $\sim$9.57\%) in person independent setting. To demonstrate that the network learned efficient features, we further trained a Support Vector Regressor (SVR) over the features learned in 31\textsuperscript{st} layer and 34\textsuperscript{th} layer for TabletGaze dataset. As depicted in TABLE~\ref{tg_compare}, the low gaze prediction errors of SVR confirms that the learned features are highly efficient.

Similarly, RAZE outperforms supervised methods~\cite{skodras2015visual,jyoti2018automatic} on CAVE dataset with 0\textdegree\ yaw angle and it is interesting to note that pre-training on `in-the-wild' data stabilizes the standard deviation significantly. Also it is quite intuitive that the eye patch based region performs the better as compared to the whole face as input. The reason being the noise introduction due to other facial parts. For experiments, we try our best to follow the protocols discussed in~\cite{skodras2015visual} and~\cite{huang2015tabletgaze}. However, there can be a few differences in frame extraction and selection.  

Similarly, we perform downstream adaptation experiments on RT-GENE and MPII datasets~\cite{zhang15_cvpr,fischer2018rt}. We use the similar evaluation protocol mentioned in~\cite{zhang15_cvpr,fischer2018rt}. The result comparison with the state-of-the-art methods are depicted in TABLE~\ref{tab:rt-gene} and~\ref{tab:mpii} respectively. We use eye patch as input for both RT-GENE and MPII dataset. For RT-GENE dataset, our self-supervised method performs better than the baseline and the state-of-the-art methods ($7.7$\textdegree\ $\rightarrow$ $6.1$\textdegree, $\sim$20.77\%). For MPII dataset, our method (angular error: 5.0\textdegree) also compatible with supervised spatial weight CNN method (angular error: 4.8\textdegree). The results on the four benchmark datasets indicate that our method learns discriminative and rich representation.

\noindent \textbf{Driver Gaze Estimation:} Another application specific downstream task is driver gaze estimation. The network is adapted for driver gaze zone estimation on DGW dataset. The hyper-parameters and other relevant details of the network is described in experiment section. We evaluate the performance of Ize-Net network by cross-validating it's performance some with other gaze estimation task. We choose Driver Gaze in the Wild (DGW)~\cite{ghosh2021speak2label} data for this purpose. It performs automatic labeling by adding domain knowledge during the data recording process and generate a large scale gaze zone estimation dataset. TABLE~\ref{tab:speak2label} shows the comparison between performance of the baseline model proposed in~\cite{ghosh2021speak2label} with Ize-Net. It is observed that our approach outperforms several supervised models with a large margin which indicates that our model learns relevant representative features.

\begin{figure}[t]
	\centering
	\subfloat{\includegraphics[width = 0.9in,height=0.5in]{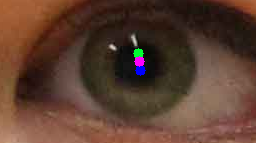}}
	\subfloat{\includegraphics[width = 0.9in,height=0.5in]{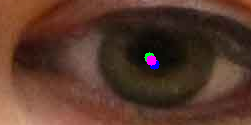}}
	\subfloat{\includegraphics[width = 0.9in,height=0.5in]{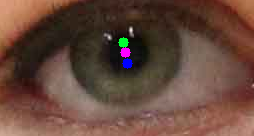}}
	\subfloat{\includegraphics[width = 0.9in,height=0.5in]{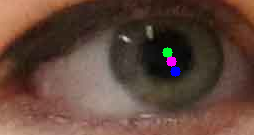}}
	\caption{Results of pupil-center localization method. Green, blue and  pink colors represent the pupil-centers as mentioned in Algorithm~\ref{alg:pupil_loc} (Image Source:~\cite{smith2013gaze} best viewed in color).} 
	\label{fig:pupil_eye}
\end{figure}

\subsection{Ablation Studies}
\subsubsection{Choice of Pupil Localization}
The pupil-center detection is performed using OTSU thresholding with blob center detection and CHT. To perform CHT, we crop the image around the pupil-center which we detect using OTSU thresholding and blob-center. We use offset of 5 pixels to crop the image. The evaluation protocol is mentioned in equation~\ref{eqn:error}, is same as the one used in~\cite{jesorsky2001robust}. 
\begin{equation}
\label{eqn:error} 
e=\dfrac{max(d_l-d_r)}{\lVert C_l-C_r \rVert}
\end{equation}
where, $e$ is the error term, $d\textsubscript{l}$ and $d\textsubscript{r}$ are the Euclidean distances between the localized pupil-centers and the ground truth ones; $C\textsubscript{l}$ and $C\textsubscript{r}$ are left and right pupil-centers respectively in the ground truth.

\noindent \textbf{Quantitative Analysis:} Table~\ref{pupil_val} shows the comparison of the proposed method with some of the state-of-the-art methods. This table shows that our method is absolutely accurate in \begin{math} e \leq 0.10 \end{math} and \begin{math} e \leq 0.25 \end{math} cases, but it does not perform well enough when \begin{math} e \leq 0.05 \end{math}. The reason behind this is the inaccurate circle detection by CHT, which propagates the error while averaging primary and secondary pupil-centers (See Algorithm~\ref{alg:pupil_loc}). 

\noindent \textbf{Qualitative Analysis:} Empirically, we observe that the pupil-center localization accuracy is increased by taking an average of pupil-centers calculated by the above two methods. Few sample results of pupil-center localization have been shown in Figure~\ref{fig:pupil_eye}. The blue, green, and pink dots represent the pupil-center obtained by our primary method, secondary method and their average, respectively.

\begin{table}[t]
	\centering
	\caption{Comparison of proposed pupil-center localization method on BioID dataset~\cite{jesorsky2001robust} with other state-of-the-art methods.}
	\label{pupil_val}
	\scalebox{1}{
		\begin{tabular}{l|c|c|c}
			\toprule[0.4mm]
			\rowcolor{mygray}
			\multirow{1}{*}{\textbf{Methods}} & \multicolumn{3}{c}{\textbf{Accuracy (\%)}}               \\ \cline{2-4} 
			\cellcolor{mygray} &\cellcolor{mygray} \textbf{\begin{math} e \leq 0.05 \end{math}} & \cellcolor{mygray} \textbf{\begin{math} e \leq 0.10 \end{math}} & \cellcolor{mygray} \textbf{\begin{math} e \leq 0.25 \end{math}} \\ \hline \hline
			\textbf{Ours}               & \textbf{56.97}    & \textbf{100.00}     & \textbf{100.00}      \\ 
			Poulopoulos et al.~\cite{poulopoulos2017new}                      & 87.10              & 98.00               & 100.00               \\ 
			Leo et al.~\cite{leo2014unsupervised}                             & 80.70              & 87.30             & 94.00                \\ 
			Campadelli et al.~\cite{campadelli2006precise}                       & 62.00                & 85.20             & 96.10              \\ 
			Cristinacce et al.~\cite{cristinacce2004multi}                     & 57.00                & 96.00               & 97.10              \\ 
			Asadifard et al.~\cite{asadifard2010automatic}                        & 47.00                & 86.00               & 96.00                \\ 
			\bottomrule[0.4mm]
	\end{tabular}}
	
\end{table}
	
\subsubsection{Choice of Gaze Heuristic}	
In order to evaluate the performance of the proposed heuristic, we compare the ground truth gaze direction derived from the CAVE dataset with the heuristic based gaze direction. The overall accuracy is approximately 87\%. The heuristic mostly fails to infer the direction when the head movement is beyond $\pm 10$\textdegree.

\subsubsection{Choice of Network Architecture}
The efficiency of the proposed eye gaze region estimation is validated on the CAVE dataset~\cite{smith2013gaze}. For this purpose, we map the angular value labels of CAVE dataset images into left, right, and central gaze regions based on the sign (positive and negative) of the gaze point. The validation results are shown in Table~\ref{result}. We also evaluate the performance of Alexnet~\cite{krizhevsky2012imagenet} and VGG-Face~\cite{parkhi2015deep} networks on the collected new dataset. AlexNet and VGG-face give 88.22\% and 84.30\% validation accuracy, respectively. We use Stochastic Gradient Descent (SGD) optimizer with categorical cross-entropy as the loss function for training both the networks. The learning rate and momentum are assigned 0.01 and 0.9 values, respectively. For quantitative anaysis, we use full face images as well as eye patch as input. From empirical analysis, it is observed that eye-patch usually performs better than full face as input. The reason behind this is that the eye patch region provide more relevant information for the gaze inference.

\subsubsection{Performance of Ize-Net Network on Pretext Task}
For training the proposed Ize-Net network, we initialize the network weights with `glorot normal' distribution. We use the SGD optimizer with a learning rate of 0.001 with the decay of $1 \times e ^{-6}$ per epoch. We use categorical cross-entropy as the loss function to train the proposed network. As mentioned in TABLE~\ref{result}, it gives 91.50\% accuracy on the validation data of the proposed dataset. The proposed network outperforms the efficiency of AlexNet and VGG-face networks. The primary reason behind the better performance of Ize-Net is the presence of the primary capsule. This enables the network to consider the geometry of the face into account during gaze region prediction. The consideration of face geometry is in accordance with the proposed heuristic used to label the collected dataset's images. We validate the performance of the proposed network on the CAVE dataset. The angular labels of CAVE dataset images have been mapped into three gaze regions. Post categorizing the images into their corresponding gaze regions, we fine-tune the Ize-Net for the entire CAVE dataset to cross-check this network's performance. We fine-tune our network for 10 epochs with 0.0001 learning rate~\cite{smith2013gaze}. As mentioned in TABLE~\ref{result}, our network gives 82.80\% five-fold cross-validation accuracy on CAVE dataset.

\begin{table}[b]
	\centering
	\caption{Validation of our proposed heuristic and Ize-Net network for CAVE dataset and proposed dataset.}
	\label{result}
	\scalebox{1}{
		\begin{tabular}{l||c|c}
			\toprule[0.4mm]
			\rowcolor{mygray}
			\textbf{Method/ Network} & \textbf{CAVE} & \textbf{\begin{tabular}[c]{@{}c@{}}RAZE\\ Dataset\end{tabular}} \\ \hline \hline
			Eye Gaze heuristic       & 60.37\%              & N/A                                                            \\ 
			Alexnet (full face)                 & N/A              & 88.22\%                                                         \\ 
			VGG-Face (full face)                 & N/A              & 84.30\%                                                         \\ 
			Ize-Net (full face)                 & 82.80\%        & 91.50\%                                                        \\ 
			Ize-Net (eye patch)                 & 88.80\%        & 95.98\%                                                      \\ 
			\bottomrule[0.4mm]
	\end{tabular}}
	
\end{table}

\begin{table}[!ht]
	\centering
	\caption{Validation results of the proposed method with voting based label smoothing.}
	\label{tab:voting}
	\begin{tabular}{l||c|c}
		\toprule[0.4mm]
		\rowcolor{mygray}
		\textbf{Method/Network}       & \textbf{CAVE} & \textbf{RAZE Dataset} \\ \hline \hline
		\textbf{Eye Gaze Heuristic}   & 62.79\%       & NA                   \\ 
		\textbf{Alexnet (Full Face)}  & NA            & 89.45\%              \\ 
		\textbf{VGG-Face (Full Face)} & NA            & 85.66\%              \\
		\textbf{Ize-Net (Full Face)}  & 81.34\%       & 90.82\%              \\ 
		\textbf{Ize-Net (Eye Patch)}  & 86.25\%       & 89.73\%              \\ 
		\bottomrule[0.4mm]
	\end{tabular}
\end{table}
	
\subsection{Voting based Label Smoothing Strategy}
We introduce label based voting in time domain (here, time domain means along the time axis of the input video) to smooth the gaze trajectory. We organize image frames in the order of appearance in the corresponding video. We select the gaze labels of five neighboring frames (in successive order) and calculate the voting over 3-zones (left, right, and central). The labels are assigned according to the max-voting strategy. The results of these experiments are shown in Table~\ref{tab:voting}. We compare the gaze estimation results with label smoothing (Table~\ref{tab:voting}) and without label smoothing (Table~\ref{result}). As compared to the gaze estimation on image frames without label smoothing, there is around 1-2\% increment in the accuracy for CAVE dataset as well as our dataset. The increment in accuracy percentage suggests that label smoothing introduced more robustness in the data labeling.

\subsection{Generalization Capability of Self-Supervised Method}

We evaluate the generalization capability of our proposed method. For this purpose, we conduct experiments by pre-training on the train part and further validate it for the downstream task of gaze estimation. We train RAZE framework on CAVE and MPII datasets to validate the performance of our self-supervised method. The results are shown in TABLE~\ref{tab:general}. The results depict the generalization capability of our proposed method.

\begin{table}[htbp]
\caption{Performance of the state-of-the-art method on CAVE and MPII datasets. * methods are supervised.}
    \label{tab:general}
    \centering
    \begin{tabular}{l||c|c|c}
    \toprule[0.4mm]
		\rowcolor{mygray}
     Methods & Pre-train   &  CAVE & MPII\\ \hline \hline
         Park et al.~\cite{park2018deep} & CAVE/MPII & 3.80\textdegree & 4.50\textdegree  \\ 
         Jyoti et al.~\cite{jyoti2018automatic}*         & CAVE &\textbf{2.22\textdegree}  & --                                                                                            \\  
                                      
                                       Yu et al.~\cite{yu2019unsupervised}         &CAVE & 3.42\textdegree    &  -- \\                             Cheng et al.~\cite{cheng2020coarse}* &MPII & -- & \textbf{4.10\textdegree}                                                         \\  
                                       
         RAZE  & CAVE/MPII &2.40\textdegree & 4.20\textdegree \\ \bottomrule[0.4mm]
    \end{tabular}
    
\end{table}

\section{Conclusion, Limitations and Future Work}
\label{sec:Conclusion and Future Work}
In this paper we propose a method for learning a rich eye gaze representation by using self-supervised learning. At first, we define the pretext task by utilizing the relative position of pupil-centers and annotate the images on three gaze region i.e. left, right, or center. To learn a rich representation, we collect a large dataset of the facial image. We also propose a capsule layer based CNN network, `Ize-Net', which is trained on the collected dataset. The learned representation is transferred into two downstream tasks. The quantitative and qualitative results indicates that the proposed method learns rich representation. 

Currently, the proposed method performs eye gaze estimation for near frontal images. We have selected the images in the dataset based on only the roll head pose angles. It is important to note here that images with varying yaw angle (within a certain range) of head pose also looks frontal. The current work does not take the variation in the yaw angle into consideration while calculating the eye gaze. Since the current approach utilizes humans' symmetrical facial features to detect the gaze-direction; the amount of error will be very less due to yaw angle variation. In the future, we plan to utilize the head pose and other relevant information while estimating the gaze region.

\section*{Acknowledgment}
We gratefully acknowledge the support of NVIDIA Corporation with the donation of the Titan Xp GPU used for this research.


\bibliographystyle{ieee}
\bibliography{ieee}

\begin{IEEEbiography}[{\includegraphics[width=1in,height=1.25in, clip,keepaspectratio]{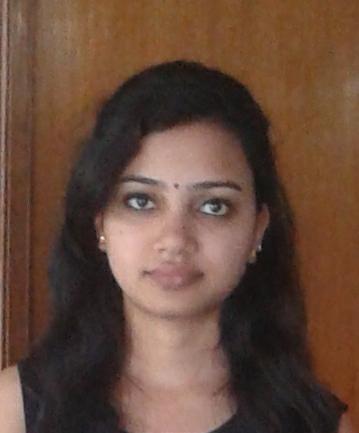}}]{Neeru Dubey} is currently a research scientist at SaleskenAI. She received her PhD from Indian Institute of Technology Ropar, India. Her research interests include computer vision, Deep Learning and Natural Language Processing. She received her bachelor's degree in Computer Science and Engineering from Guru Gobind Singh Indraprastha University (New Delhi, India). Her research interest is HCI, Computer Vision and AI.
	
\end{IEEEbiography}

\begin{IEEEbiography}[{\includegraphics[width=1in,height=1.25in,clip,keepaspectratio]{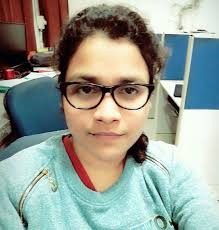}}]{Shreya Ghosh} is currently pursuing PosDoc at Monash University. Her PostDoc is funded by DARPA. She is on the way to complete her PhD from Monash University, Australia. She received MS(R) degree in the Computer Science and Engineering from the Indian Institute of Technology Ropar, India. She received the bachelor’s degree in Computer Science and Engineering in 2016 from the Govt. College of Engineering and Textile Technology Serampore (West-Bengal, India). Her research interests include Affective computing, computer vision, Deep Learning. She is a student member of the IEEE.
	
\end{IEEEbiography}

\begin{IEEEbiography}[{\includegraphics[width=1in,height=1.25in,clip,keepaspectratio]{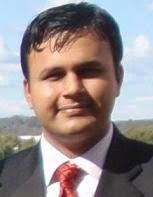}}]{Abhinav Dhall} is an Assistant Professor at Indian Institute of Technology Ropar and Adjunct Senior Lecturer at Monash University. He received PhD from the Australian National University in 2014. Followed by postdocs at the University of Waterloo and the University of Canberra. He was awarded the Best Doctoral Paper Award at ACM ICMR 2013, Best Student Paper Honourable mention at IEEE AFGR 2013 and Best Paper Nomination at IEEE ICME 2012. His research interests are in computer vision for Affective computing and Assistive Technology. He is a member of the IEEE and Associate Editor of IEEE Transactions on Affective Computing.
\end{IEEEbiography}
	
\end{document}